\documentclass{article}
\usepackage{spconf,amsmath,graphicx}

\usepackage{booktabs}
\usepackage{url}
\usepackage{subcaption}
\usepackage{xcolor}
\usepackage{hyperref}
\hypersetup{
    colorlinks,
    linkcolor={blue!50!black},
    citecolor={blue!50!black},
    urlcolor={blue!50!black}
}
\usepackage{enumitem}
\setlist{nosep, leftmargin=14pt}

\title{Combining imaging and shape features for prediction tasks of Alzheimer's disease classification and brain age regression}
\name{Nairouz Shehata $\,^{1,2}$
\qquad Carolina Piçarra $\,^1$
\qquad Ben Glocker $\,^1$}
\address{$^1$ Imperial College London, UK \quad $^2$ Aswan Heart Centre, Egypt}
\begin{document}

\maketitle

\begin{abstract} 
We investigate combining imaging and shape features extracted from MRI for the clinically relevant tasks of brain age prediction and Alzheimer's disease classification. Our proposed model fuses ResNet-extracted image embeddings with shape embeddings from a bespoke graph neural network. The shape embeddings are derived from surface meshes of 15 brain structures, capturing detailed geometric information. Combined with the appearance features from T1-weighted images, we observe improvements in the prediction performance on both tasks, with substantial gains for classification. We evaluate the model using public datasets, including CamCAN, IXI, and OASIS3, demonstrating the effectiveness of fusing imaging and shape features for brain analysis.
\end{abstract}

\section{Introduction}

Alzheimer's disease (AD) is a progressive neurodegenerative disorder, leading to substantial memory impairment, making early diagnosis crucial \cite{buchake2010neurodegenerative}. Additionally, predicting brain age using MRI has become a clinically important method for identifying imaging biomarkers associated with various neuropathologies. Many studies have proposed brain age prediction as a means to characterize these conditions \cite{cole2017predicting}. 

Neuroimaging, such as structural MRI, has become indispensable for diagnosis. With the availability of large-scale datasets, there has been a surge in applying sophisticated machine learning algorithms to enhance our understanding and prediction capabilities of neurodegenerative disease \cite{gao2022review,menagadevi2024machine}. The complex anatomy of brain structures necessitates sophisticated approaches to fully capture spatial relationships and geometric properties. The advancement of automated segmentation techniques \cite{menagadevi2024machine}, facilitates anatomical shape extraction from medical images. Extracted shapes can then be represented as meshes, enabling the use of powerful Graph Neural Networks (GNNs) for learning geometric features. 

In this paper, we present a deep learning approach that integrates imaging features with brain shape features extracted from structural MRI to enhance prediction accuracy for brain age regression and AD classification. Our approach takes advantage of the strengths of both image-based and graph-based deep learning techniques. Specifically, we employ a ResNet-18 architecture for extracting features from the T1-weighted MR images and a multi-graph neural network architecture for processing brain surface meshes. By combining these complementary feature embeddings, our model aims to capture a comprehensive representation of brain anatomy, potentially improving predictive performance compared to models using either imaging or shape features alone. Our results demonstrate that the fusion model outperforms baseline models, highlighting the potential of combining image-based and graph-based features for more accurate predictions.

\section{Methods}
\subsection{Imaging datasets} 
We use three public datasets; mCAN\cite{taylor2017cambridge,shafto2014cambridge}, IXI \footnote{\url{https://brain-development.org/ixi-dataset/}}, and OASIS3 \cite{lamontagne2019oasis}. Images in CamCAN were acquired using a 3T scanner. IXI includes data from three clinical sites, with 3T and 1.5T scanners. OASIS3 comprises 716 subjects with normal cognitive function and 318 patients with various stages of cognitive decline. Data with missing information on biological sex or age were excluded. Pre-processing included skull stripping with ROBEX \cite{iglesias2011robust}, bias field correction using N4ITK \cite{tustison2010n4itk}.The 15 subcortical brain structures\footnote{brain stem, left and right thalamus, caudate, putamen, pallidum, hippocampus, amygdala, and accumbens-area} were automatically segmented as triangular meshes FSL FIRST \cite{patenaude2011bayesian}.

\subsection{Brain shape representation} 
Brain meshes are undirected graphs where nodes store additional information represented as feature vectors. We employ Fast Point Feature Histograms (FPFH) \cite{rusu2009fast}, a powerful feature descriptor as demonstrated in \cite{shehata2022comparative}. Calculating FPFH featurescludes approximating normals and computing angular differences, leading to a 33-characteristic feature vector.

All meshes are aligned to a structure-specific standard orientation by performing rigid registration using the closed-form Umeyama method \cite{umeyama1991least}. One subject is selected randomly from the dataset serving as the reference mesh.

\subsection{Feature extraction \& fusion} 
\begin{figure}
    \centering
    \includegraphics[width=\columnwidth, trim = 9cm 5cm 5cm 3cm, clip]{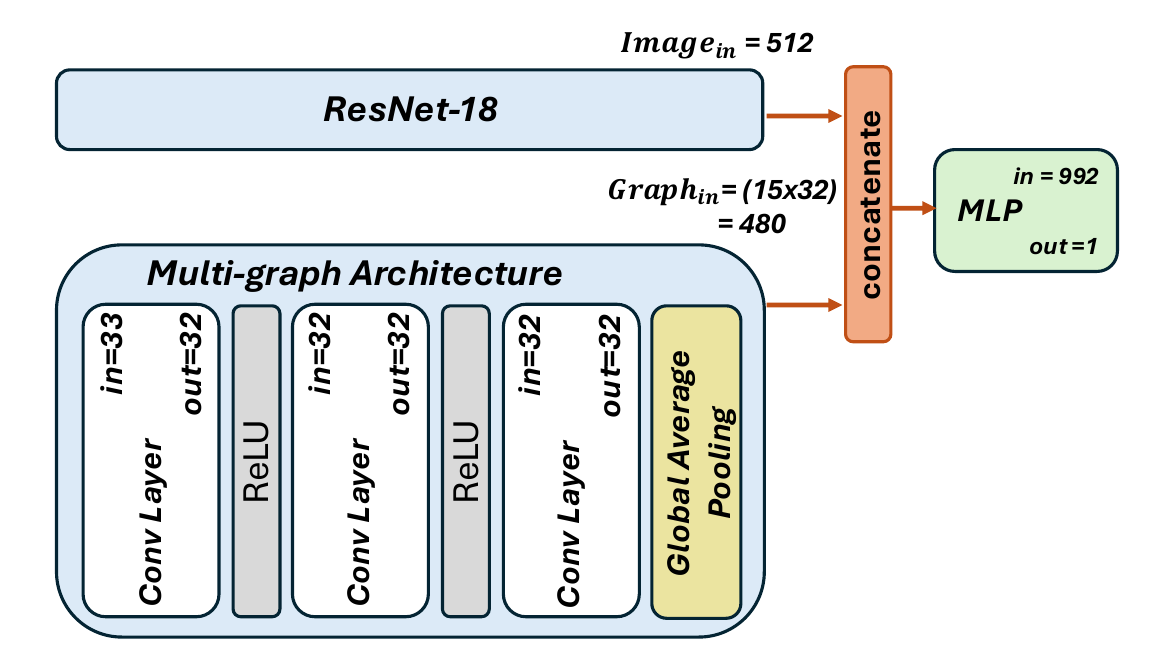}
    \caption{\textbf{Fusion model overview}, consisting of image-based and graph-based feature extractors, embeddings concatenated and passed through a multi-layer perceptron (MLP).
    }
    \label{fig:fusion_model}
\end{figure}

Our feature extraction process employs two distinct models; an image-based model using ResNet-18 architecture and a graph-based model utilizing a bespoke multi-graph architecture described below. Figure \ref{fig:fusion_model} provides an overview of our fusion model, illustrating how the two feature extraction models are integrated. The readouts from the pooling layer of the graph-based model and from the second to last layer of the ResNet are concatenated and passed to a linear predictor. 

\textbf{Imaging features.} We use ResNet-18 from PyTorch \cite{paszke2019pytorch}, with the last linear layer removed after pre-training. Data augmentation consists of horizontal flip, gamma intensity adjustment, addition of Gaussian noise with random parameters, and application of a random affine transformation.

\textbf{Shape features.} To utilise all 15 brain shapes, a multi-graph architecture is designed with a submodel for each shape. The embeddings are concatenated and fed into two FC layers. The submodels have three convolutional layers with Rectified Linear Unit (ReLU) activations and a global average pooling layer to aggregate shape representations \cite{shehata2022comparative}. Using structure-specific submodels was previously found to yield more discriminative shape features \cite{shehata2024importance}. 

\section{Experiments \& Results}
The same samples per data split in training, validation, and test sets were used consistently for the image-based, graph-based models, and fusion models ensuring a fair comparison between all approaches. Implementations of the models and data handling are based on PyTorch Geometric \footnote{\url{https://pytorch-geometric.readthedocs.io/}} and PyTorch Lightning \footnote{\url{https://lightning.ai/docs/pytorch/stable/}}. We used the Adam optimiser with a learning rate of 0.001 and employed the standard cross-entropy loss for AD classification and mean squared error for brain age regression.

Random node translations with a maximum strength of 0.1mm was applied as data augmentation when training the graph-based models \cite{zhou2020data}. For the graph convolutional layers, we compare SplineConv \cite{fey2018splinecnn} and GCNConv \cite{kipf2016semi}.

We extract the feature embeddings from three different models; the image-based model (ResNet) and two graph-based models (SplineConv and GCNConv). The embeddings are passed through an MLP independently or concatenated for the fusion models. Three random seeds were used to calculate the mean and standard deviation of the performance.

\subsection{Brain age regression} 
\begin{figure*}
\centering
\begin{subfigure}{0.16\linewidth}
\centering
\includegraphics[width=1.1\linewidth]{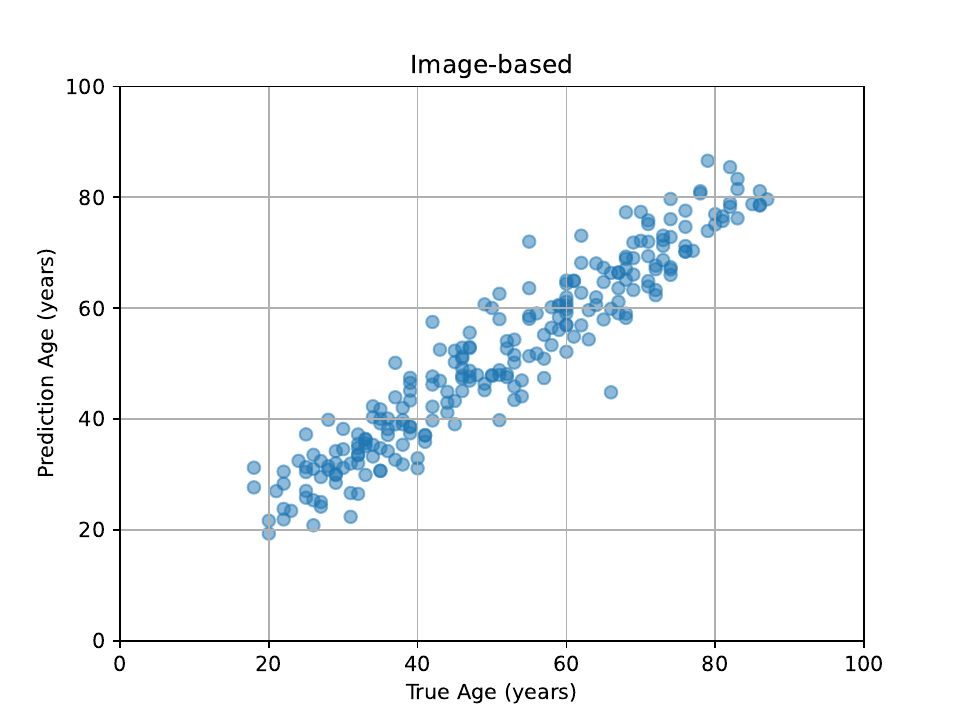}
\caption{ResNet}
\end{subfigure}
\qquad
\begin{subfigure}{0.16\linewidth}
\centering
\includegraphics[width=1.1\linewidth]{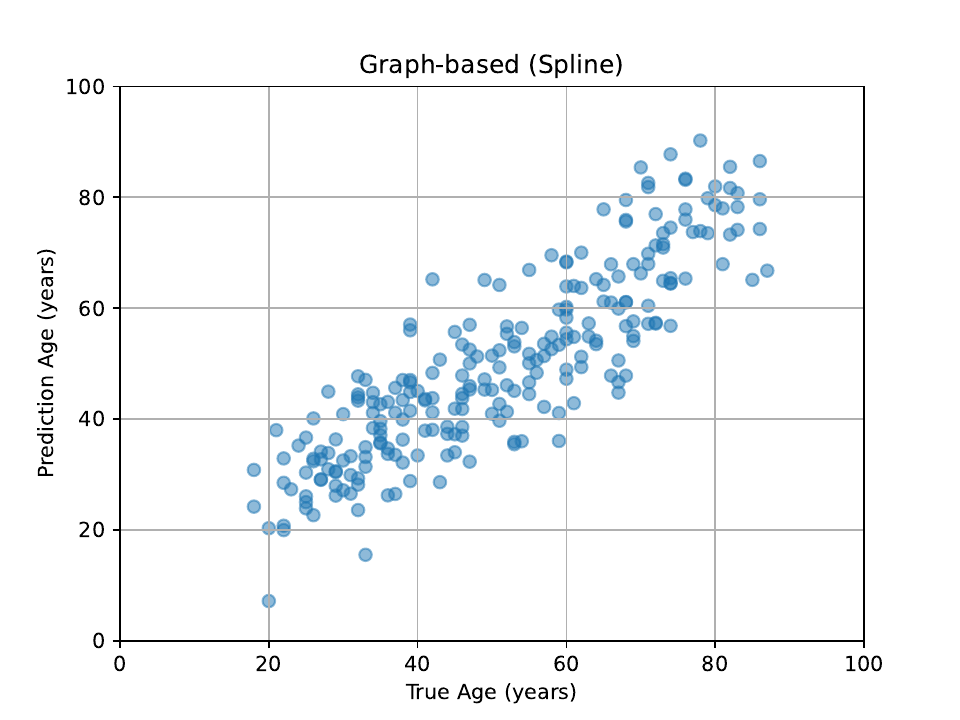}
\caption{SplineConv}
\end{subfigure}
\qquad
\begin{subfigure}{0.16\linewidth}
\centering
\includegraphics[width=1.1\linewidth]{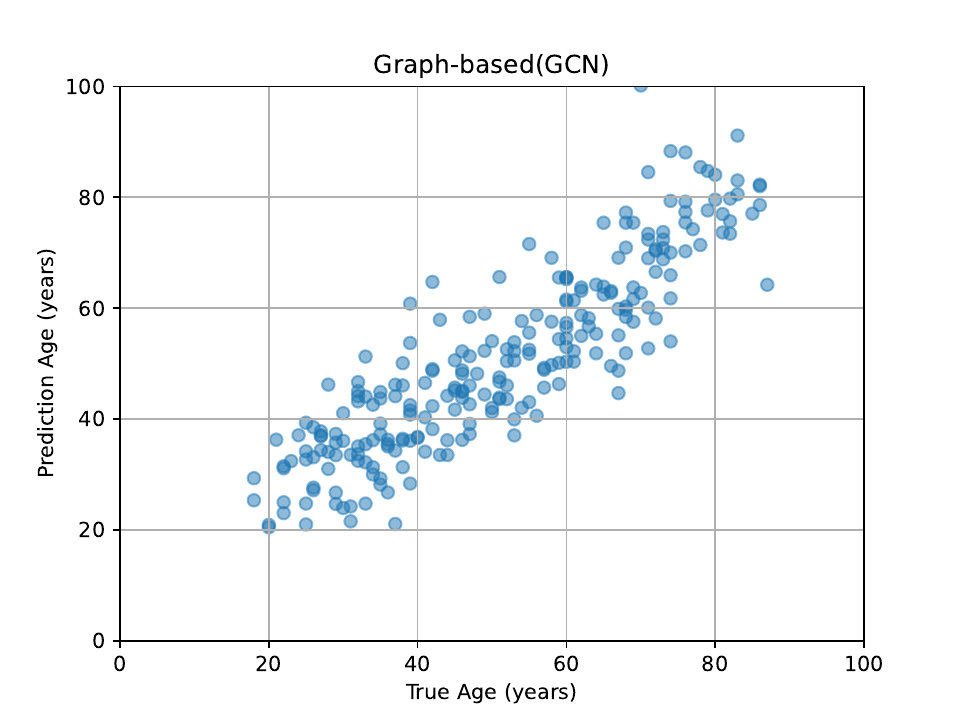}
\caption{GCNConv}
\end{subfigure}
\qquad
\begin{subfigure}{0.16\linewidth}
\centering
\includegraphics[width=1.1\linewidth]{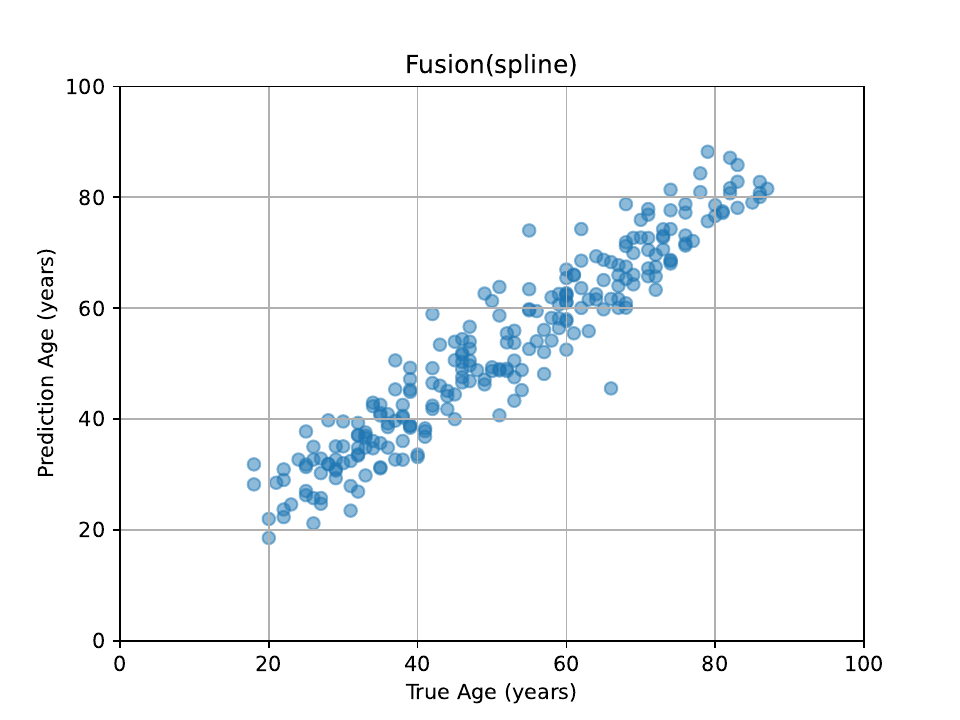}
\caption{Fusion SplineConv}
\end{subfigure}
\qquad
\begin{subfigure}{0.16\linewidth}
\centering
\includegraphics[width=1.1\linewidth]{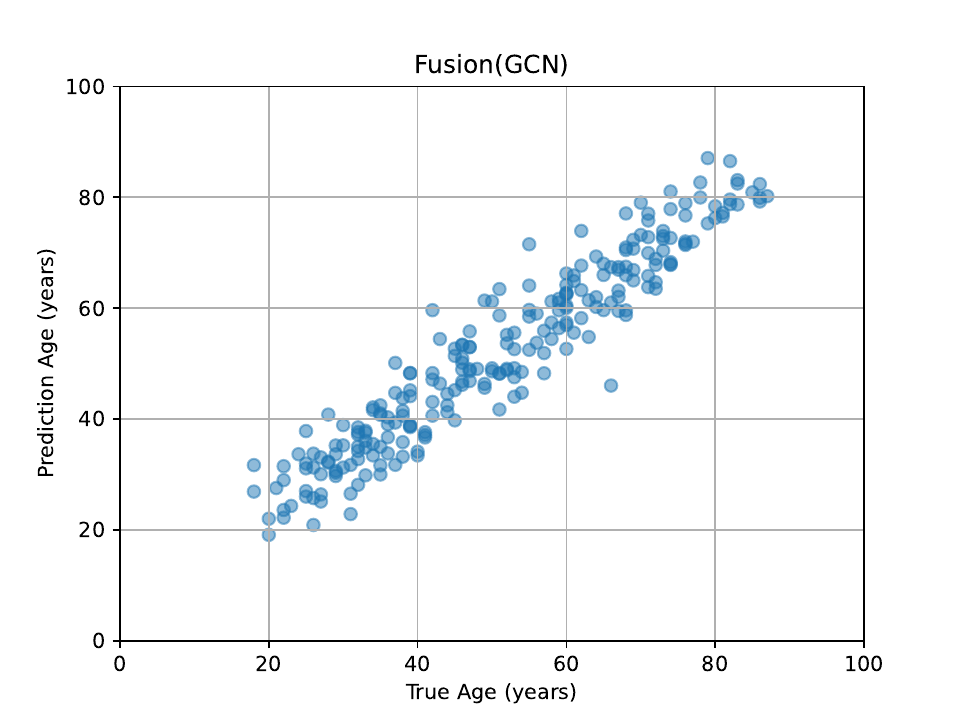}
\caption{Fusion GCNConv}
\end{subfigure}
\caption{Scatter plots of true age against predicted age for different brain age prediction models.}
    \label{fig:age_predictions}
\end{figure*}

We split the combined dataset of CamCAN and IXI into 875 training, 97 validation and 243 testing samples. The age range is 18-88 years. The models are evaluated using mean absolute error (MAE), %calculating the average difference 
and R2 scores. Table \ref{tab:age_resnet_graph_1fc} shows the results over three runs for five different models. The scatter plots in Fig. \ref{fig:age_predictions} visualise the relationship between true and predicted ages.

\begin{table}[h!]
    \centering
        \begin{tabular}{lcc}
            \toprule
            Model & MAE  $\downarrow$ & R2 score  $\uparrow$ \\
            \midrule
            ResNet & 4.391±0.005 & 0.889±0.000 \\
            % 4.39143±0.004927 0.88863±4.71e-05
            SplineConv & 7.151±0.027 & 0.724±0.007\\ 
            % 7.151±0.02695& 0.72449±0.00676
            GCNConv & 6.734±0.035 & 0.741±0.005\\
            \midrule
            Fusion w/ SplineConv & \textbf{4.362±0.074} & \textbf{0.893±0.004} \\
            % 4.3623±0.07387 0.89283±0.00366
            Fusion w/ GCNConv & 4.388±0.037 & 0.890±0.001\\
            % 4.388233333333333 0.03739272062254304

            \bottomrule
        \end{tabular}
    \caption{Evaluation metrics for brain age regression using embeddings extracted from image- and graph-based models.}
    \label{tab:age_resnet_graph_1fc}
\end{table}

\subsection{Alzheimer's disease classification} 
\begin{figure}
    \centering
    \includegraphics[width=0.37\textwidth]{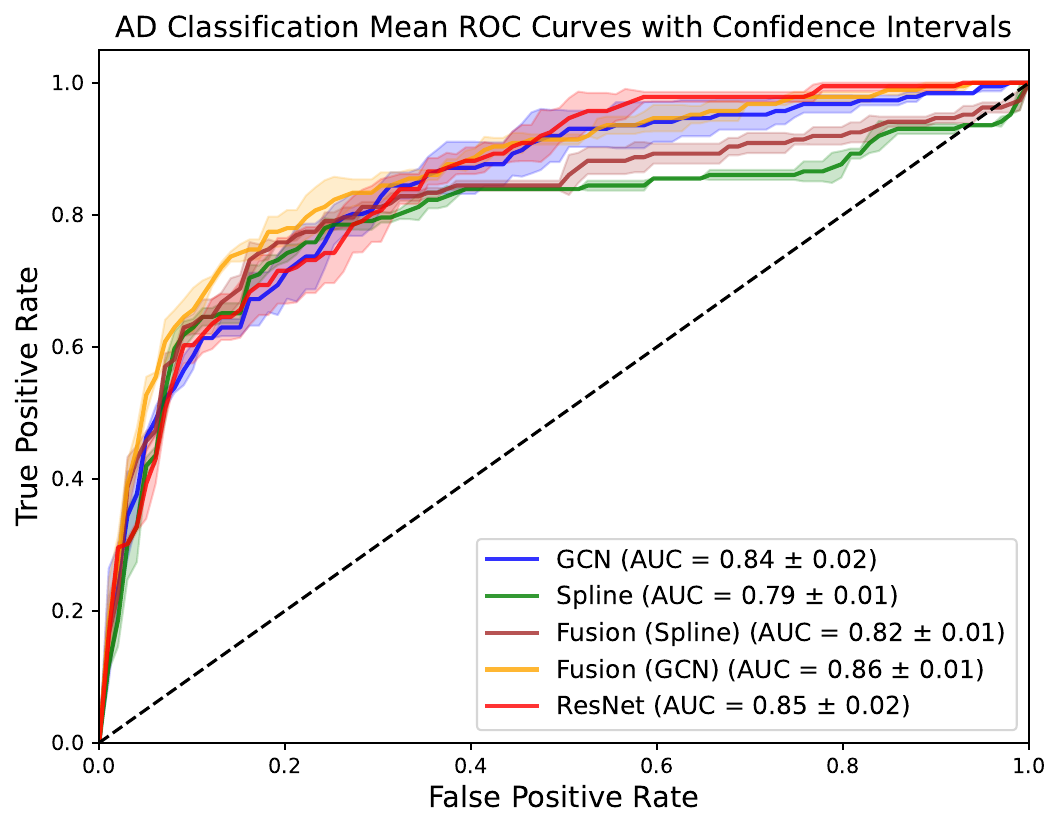}
    \caption{ROC curves reveal substantial performance improvements for fusion models in the low FPR regime.}
    \label{fig:1fc_ad_ROC}
\end{figure}

OASIS-3 was split into 745 training, 82 validation, and 207 testing samples, with 62 positives in the test sample. The models are evaluated using the area under the ROC curve (AUC) and true and false positive rates. We report TPRs at fixed FPR of 0.15 and 0.20, and FPR at fixed TPR of 0.7. Classification results are shown in Table \ref{tab:AD_standalone_1FC} for the five models, over three runs. The ROC curves are shown in Fig. \ref{fig:1fc_ad_ROC}.

\begin{table*}
    \centering
    \resizebox{\textwidth}{!}{
        \begin{tabular}{lcccc}
            \toprule
            Model & AUC $\uparrow$ & TPR@FPR=0.15  $\uparrow$ & TPR@FPR=0.20 $\uparrow$ & FPR@TPR=0.70  $\downarrow$ \\
            \midrule
            ResNet & 0.848 ± 0.018& 0.656 ± 0.042&  0.715 ± 0.050& 0.172 ± 0.046\\
            SplineConv & 0.791 ± 0.005& 0.651 ± 0.015& 0.742 ± 0.013& 0.162 ± 0.020\\
            GCNConv & 0.837 ± 0.019& 0.629 ± 0.013&  0.715 ± 0.042& 0.192 ± 0.046\\
            \midrule
            Fusion w/ SplineConv & 0.817 ± 0.008& 0.688 ± 0.020& 0.758 ± 0.000& 0.152 ± 0.020\\
            Fusion w/ GCNConv & \textbf{0.861 ± 0.007}& \textbf{0.742 ± 0.013} & \textbf{0.780 ± 0.027} & \textbf{0.121 ± 0.008} \\
            \bottomrule
        \end{tabular}
    }
    \caption{AD classification results from using embeddings from image- and graph-based models compared to fusing embeddings from both. We report the AUC, TPR@FPR and FPR@TPR at varying performance levels. We observe that using fused feature embeddings outperforms standalone models consistently across TPR/FPR performance levels.}
    \label{tab:AD_standalone_1FC}
\end{table*}

% GCN - TPR at FPR 0.15: 0.629 ± 0.013
% GCN - TPR at FPR 0.2: 0.715 ± 0.042
% GCN - FPR at TPR 0.7: 0.192 ± 0.046
% Spline - TPR at FPR 0.15: 0.651 ± 0.015
% Spline - TPR at FPR 0.2: 0.742 ± 0.013
% Spline - FPR at TPR 0.7: 0.162 ± 0.020
% Fusion (Spline) - TPR at FPR 0.15: 0.688 ± 0.020
% Fusion (Spline) - TPR at FPR 0.2: 0.758 ± 0.000
% Fusion (Spline) - FPR at TPR 0.7: 0.152 ± 0.020
% Fusion (GCN) - TPR at FPR 0.15: 0.742 ± 0.013
% Fusion (GCN) - TPR at FPR 0.2: 0.780 ± 0.027
% Fusion (GCN) - FPR at TPR 0.7: 0.121 ± 0.008
% ResNet - TPR at FPR 0.15: 0.656 ± 0.042
% ResNet - TPR at FPR 0.2: 0.715 ± 0.050
% ResNet - FPR at TPR 0.7: 0.172 ± 0.046

\section{Discussion}
For brain age regression, we observe some improvement with fusion models, but it remains unclear whether fusion truly adds significant value over standalone models. The graph-based models which extract shape features may offer complementary information to the image-based model, but their impact is limited when combined for age prediction (cf.  Table \ref{tab:age_resnet_graph_1fc}). While in standalone, we observe better performance for GCNConv, a slight improvement is observed for the fusion model when using SplineConv. This observation suggests that better standalone performance may not always translate to better performance in a fusion model.

In contrast, we observe a substantial improvement with fusion models for AD classification, suggesting that the fusion of imaging and shape features is beneficial for this more challenging classification task. The results in Table \ref{tab:AD_standalone_1FC} indicate that combining the features leads to better performance in the critical regions of lower FPR, which is particularly important for clinical applications. Interestingly, using AUC alone does not fully reveal these improvements. While AUC values for the ResNet and fusion models appear similar (0.848 vs 0.861), a closer inspection of the ROC curves shows noticeable performance differences in the clinically relevant areas of low FPR. Additionally, by evaluating TPR at fixed FPR values and FPR at fixed TPR, we see substantial improvements in AD classification performance using fusion models. Emphasizing that these metrics provide a more detailed understanding of the model's behaviour in the regions of the ROC curve that are most important for practical use. This demonstrates that the fusion of imaging and shape features consistently leads to more reliable AD classification.

\section{Conclusion}

By leveraging deep learning architectures tailored for different data representations, images and shapes, we find substantially improved predictive accuracy for AD classification. Importantly, evaluating models for AD classification using ROC curves, rather than AUC alone, reveals performance differences at clinically relevant classification thresholds.

Future work will focus on evaluating the model on larger and more diverse datasets and other prediction tasks. Additionally, we plan to explore the use of other image modalities, such as functional MRI, to further enhance the model's performance. While our simple approach of concatenating imaging and shape features showed improvements, more sophisticated fusion methods, such as attention mechanisms or adaptive fusion, could be beneficial. We also plan to interpret the contribution of each modality to the final predictions. This could provide valuable insights into the structural changes associated with brain ageing and AD progression.

This work may also inspire the use of different data representations in other domains, outside neuroimaging. Prediction tasks such as malignancy of lung nodules could benefit from the fusion of imaging and shape features.

\section{Acknowledgments}
\label{sec:acknowledgments}

Nairouz Shehata is grateful for the support of the Magdi Yacoub Heart Foundation and Al Alfi Foundation.

\section{Compliance with ethical standards}
\label{sec:ethics}

This study uses secondary, fully anonymised data which is publicly available and is exempt from ethical approval.

\bibliographystyle{IEEEbib}
\bibliography{strings,refs}

\end{document}